\documentclass[sigconf]{acmart}
\pdfoutput=1

\usepackage{graphicx}
\usepackage{booktabs, tabularx, array, multirow, colortbl, xcolor}
\usepackage{adjustbox}
\usepackage{caption, subcaption}
\usepackage{float} 
\usepackage{dblfloatfix}
\usepackage{algorithm, algorithmic}
\usepackage{enumitem, xspace, makecell}
\usepackage{pgfplots}
\pgfplotsset{compat=1.18}
\usepackage{bp-listings}

\usepackage{hyperref}
\title{A Behavior-Based Knowledge Representation Improves Prediction of Players' Moves in Chess by 25\%}

\author{Benny Skidanov}
\affiliation{
  \institution{Ben-Gurion University}
  \city{Be'er Sheva}
  \country{Israel}}
\email{skidanov@post.bgu.ac.il}

\author{Daniel Erbesfeld}
\affiliation{
  \institution{Ben-Gurion University}
  \city{Be'er Sheva}
  \country{Israel}}
\email{erbesfel@post.bgu.ac.il}

\author{Gera Weiss}
\affiliation{
  \institution{Ben-Gurion University}
  \city{Be'er Sheva}
  \country{Israel}}
\email{geraw@bgu.ac.il}

\author{Achiya Elyasaf}
\affiliation{
  \institution{Ben-Gurion University}
  \city{Be'er Sheva}
  \country{Israel}}
\email{achiya@bgu.ac.il}

\begin{document}

\setcopyright{acmlicensed}
\copyrightyear{2025}
\acmYear{2025}

\begin{abstract}
Predicting player behavior in strategic games, especially complex ones like chess, presents a significant challenge. The difficulty arises from several factors. First, the sheer number of potential outcomes stemming from even a single position, starting from the initial setup, makes forecasting a player's next move incredibly complex. Second, and perhaps even more challenging, is the inherent unpredictability of human behavior. Unlike the optimized play of engines, humans introduce a layer of variability due to differing playing styles and decision-making processes. Each player approaches the game with a unique blend of strategic thinking, tactical awareness, and psychological tendencies, leading to diverse and often unexpected actions. This stylistic variation, combined with the capacity for creativity and even irrational moves, makes predicting human play difficult. Chess, a longstanding benchmark of artificial intelligence research, has seen significant advancements in tools and automation. Engines like Deep Blue, AlphaZero, and Stockfish can defeat even the most skilled human players. However, despite their exceptional ability to outplay top-level grandmasters, predicting the moves of non-grandmaster players, who comprise most of the global chess community---remains complicated for these engines.
This paper proposes a novel approach combining expert knowledge with machine learning techniques to predict human players' next moves. By applying feature engineering grounded in domain expertise, we seek to uncover the patterns in the moves of intermediate-level chess players, particularly during the opening phase of the game. Our methodology offers a promising framework for anticipating human behavior, advancing both the fields of AI and human-computer interaction.
\end{abstract}



\maketitle

\section{Introduction}
The challenge of predicting human actions is a complex and fascinating problem. Understanding and anticipating how humans make decisions requires a deep comprehension of the underlying thought processes and strategies they employ. To predict human actions effectively, it is essential to model the concepts and considerations that individuals consider while forming their strategies. In chess, this means modeling the guidelines and principles outlined in chess textbooks. These guidelines provide a foundation for understanding how players evaluate positions and decide their moves. As we will show, this type of knowledge is crucial for capturing the essence of human decision-making.

This paper introduces a novel approach using Behavioral Programming (BP)~\cite{harel2012behavioral, elyasaf2021cobp} to model chess strategies. BP enables us to represent each strategic guideline as a distinct scenario and anti-scenario, directly aligning our model with established chess principles (see \autoref{sec:BP}). This granular approach allows us to capture the intricate details of human strategic thought in chess. We extract game features from the BP model, representing the dynamic state of each strategy during gameplay, and then employ machine learning (ML) to predict human player actions. Our results demonstrate that this BP-driven modeling of chess strategies significantly improves move prediction, even with relatively simple ML algorithms, outperforming state-of-the-art, deep learning-based approaches like Maia Chess (see \autoref{sec:maia}), which rely on massive datasets and lack explicit knowledge representation. Furthermore, the  ML models we generate are more transparent and interpretable, offering valuable insights into the human decision-making process in chess. By leveraging BP, we achieve a deeper understanding and more accurate prediction of human moves, effectively bridging the gap between human intuition and machine learning.  Our approach, using simpler, lighter, and more analyzable ML models, achieves superior performance compared to Maia, highlighting the power of incorporating explicit behavioral knowledge representation into models of human decision-making.

While incorporating expert knowledge is expected to lead to some improved accuracy, the dramatic performance gains we observe, coupled with a significant reduction in computational resources (i.e., using ML instead of deep learning, eliminating the need for GPUs, and achieving faster runtime), underscore the effectiveness of our approach in capturing how human players actually make decisions. This contribution is twofold:  we achieve a substantial improvement over state-of-the-art methods, and we introduce a novel, modular approach for modeling human decision-making processes for machine learning. It is important to acknowledge, however, that this improved performance and efficiency come with a cost. Our approach requires the involvement of a domain expert and a time investment for the knowledge modeling phase. Nevertheless, our results show that the substantial gains in accuracy and the dramatic reduction in resource requirements make this investment worthwhile.


\section{Baseline --- Maia Chess}
\label{sec:maia}
Maia Chess~\cite{mcilroyyoung2020maia, tang2024maia} represents the current state-of-the-art for predicting human chess moves, and we use it as the baseline for evaluating our approach. Maia is a neural network derived from the AlphaZero framework\cite{silver2018general}, trained on human-played games without tree search, specifically designed to predict human move choices. A key distinction between Maia and our method is that Maia does not explicitly model the cognitive processes that inform human decision-making. Maia's methodology includes the following key elements:
\begin{itemize}
    \item \textit{Training on Human Games:} Trained on a vast dataset of human games from Lichess.org.
    \item \textit{Move Prediction Without Tree Search:} Relies solely on the policy network for human-like move prediction.
    \item \textit{Binary Classification Task:} Identifies blunders (a critically bad mistake) using a custom deep residual neural network.
\end{itemize}

Both Maia and our approach aim to model human decision-making in chess but differ in methodology.

\textbf{Similarities:}
\begin{itemize}
    \item \textit{Human-Centric Modeling:} Both align AI predictions with human behavior.
    \item \textit{Skill-Level Considerations:} Both recognize varying cognitive patterns across skill levels.
\end{itemize}

\textbf{Differences:}
\begin{itemize}
    \item \textit{Data Processing:} Maia uses raw gameplay data while we preprocess data through a behavioral programming simulator. As a result, Maia's features are based on the board state (i.e., the location of each piece), while our features are based on the state of each game strategy.
    \item \textit{Prediction Tasks:} Maia focuses on move prediction and blunder classification; our framework integrates strategic modeling with classical AI models.
    \item \textit{Architectural Design:} Maia uses neural networks; our approach relies on machine learning. 
\end{itemize}

Maia achieves high move-matching accuracy, outperforming engines like Stockfish~\cite{Stockfish} and Leela Chess Zero~\cite{leela}, designed to find the best move rather than mimic human behavior. Our method combines simulator-tuned features with classical models, achieving competitive predictive accuracy. 

\section{Related Work}
\label{sec:related-work}

This section reviews existing research on predicting human actions in chess.
Maia Chess has emerged as the current state-of-the-art framework among the various approaches explored in the literature. Given its prominence and relevance to our research objectives, we have exclusively compared our proposed methodology against Maia Chess. 

Predicting human chess moves has been a longstanding challenge in artificial intelligence (AI) research. Various methodologies have been explored, ranging from deep learning techniques to explicit knowledge representation and behavioral modeling. This section reviews key contributions in the field, categorizing them based on their approach to chess move prediction.

By leveraging large-scale game datasets, deep learning methods have been widely used to predict chess moves. Convolutional Neural Networks (CNNs) have been particularly effective in learning spatial patterns from chessboard images~\cite{attlstm_chess, cnn_moves}. Recurrent architectures, such as LSTMs with attention mechanisms, have also been employed to capture sequential dependencies in gameplay~\cite{chess_moves_prediction}. These methods achieve high accuracy but require extensive computational resources and lack interpretability.

A study by Luangamornlert and Theeramunkong~\cite{chess_sequential_data} examined various chessboard representations for outcome prediction, demonstrating that image-based inputs improve performance. Similarly, Panchal et al.~\cite{chess_moves_prediction} found that deep neural networks outperformed traditional approaches in move prediction but at higher computational demands. Their models achieved an accuracy of 56.15\% for knight moves but struggled with long-range pieces such as rooks and queens, which had lower accuracies of 29.25\% and 26.52\%, respectively.

Several studies have explored player-specific models that account for individual decision-making tendencies. McIlroy-Young et al.~\cite{mcilroy2022learning} proposed reinforcement learning models trained on player-specific datasets, achieving improved move prediction accuracy. Zhang et al.~\cite{personalized_behavior} further demonstrated a 12\% improvement in accuracy when personalizing models to specific players.

Another study~\cite{chess_memory} examined how chess knowledge affects memory retention and position recall accuracy. Participants reconstructed structured and random chess positions, with skilled players demonstrating superior recall of structured positions. This suggests pattern recognition aids decision-making and may make skilled players more predictable due to adherence to known strategic patterns.

A hybrid approach integrating behavioral modeling with AI has also been explored. Khan et al.~\cite{chess_attention} combined eye-tracking data with CNN-based models, improving move prediction accuracy by 15\% by incorporating attention-based insights.

While deep learning methods such as CNN and LSTM have achieved notable success in move prediction, they lack interpretability and require significant computational resources. Personalized models enhance accuracy by tailoring predictions to individual players, though they require careful consideration of data privacy and ethical concerns. Additionally, personalized models may struggle to generalize well to new or unseen data, leading to potential overfitting issues.

Our approach leverages Behavioral Programming to explicitly model generic playbook chess strategies and incorporate them into machine learning algorithms. This provides a more interpretable alternative to deep learning methods. By integrating expert knowledge into machine learning models, we aim to bridge the gap between computational efficiency and predictive accuracy, enhancing move prediction for intermediate-level players.

\section{Preliminary: A very short introduction to BP}
\label{sec:BP}
Behavioral Programming (BP) is a modeling approach where short code fragments, called b-thread, written in JavaScript, operate on a request, watch, and block protocol to coordinate event selection dynamically. Each behavioral thread (b-thread) can request events it wants to trigger, watch events it needs to monitor, and block events it wishes to prevent. At each execution step, a central event selection mechanism chooses a requested but not blocked event, ensuring synchronization among b-threads. This enables modular and incremental specification of behaviors, where different aspects can independently contribute to system execution while resolving conflicts dynamically. See~\cite{Harel2010ProgrammingCoordinated} for more details on BP. See \autoref{sec:method:bp} for a code example.

\section{Our Approach for Predicting Human Moves}
\label{sec:outline}
This section outlines our methodology for predicting chess moves using machine learning. The process involves the following key steps:
\begin{enumerate}
    \item Read the textbook and extract behavioral aspects (e.g., strategies).
    \item Use BP to model the game as a composition of these aspects.
    \item Construct a gameplay dataset where each row contains the state of the BP program before and after a move.
    \item Train a standard ML model to predict moves.
\end{enumerate}

Note that the dataset captures the state of the BP program before and after each move, not the entire state of the game. While the state of the BP program provides the necessary information for the model to learn the game's dynamics, it does not include all the information available in the whole game state. This simplification is intentional, as it allows us to focus on the behavioral aspects of the game that are most relevant for predicting human players' moves. This is similar to the use of encoders in reinforcement learning, where the state representation is designed to capture the essential features of the environment that are relevant for decision-making.

We now detail each methodology step, providing insights into the process.

\subsection{Extracting Behavioral Aspects}
The first step involves studying the fundamental strategies and heuristics players use to make decisions during a game. This can be achieved by analyzing a comprehensive chess textbook, which encapsulates gameplay's core principles and tactics. While capturing every nuance of human decision-making may not be feasible, focusing on the primary considerations and common patterns can provide a solid foundation for modeling player behavior. Our research focuses on game openings; thus, we defined only aspects that are relevant to openings. As we now elaborate, these aspects include five counters, eight basic strategies, and three advanced strategies.

\subsubsection*{Counters Specification}
One of the simplest yet fundamental aspects analyzed by our chess behavioral threads involves tracking the movements of individual pieces and correlating these movements with specific counters. Each move a piece makes increments its corresponding counters, a relationship easily derived from the game's data encoded in the common Portable Game Notation (PGN). 

We count Pawn, Knight, Bishop, Rook, and Queen moves, as these pieces are the primary actors in the opening phase of the game. These counters hold significant analytical value, capturing essential patterns of play that can reveal various strategic insights. 
Specifically, the counters for pawn and minor piece (knight and bishop) movements reflect well-known opening strategies in chess. A higher number of moves made by pawns and minor pieces can indicate the effective execution of two key principles: controlling the central squares and developing pieces early in the game. High counters for these pieces' moves are expected from intermediate players at the beginning of the game, as these principles are emphasized in the textbooks.

Conversely, strategic guidelines in chess advise against excessive movement of the queen and rooks during the opening. Best practices suggest that the queen should ideally not be moved more than a few times, as frequent movement can lead to vulnerabilities and loss of tempo. Additionally, moving rooks early may hinder the player's ability to castle, an essential maneuver for king safety. As such, lower movement counts for these pieces are generally anticipated and can indicate adherence to established opening principles.

\subsubsection*{Basic Strategies Specification}
Here, we outline several basic strategies incorporated into our behavioral threads. These strategies are fundamental to chess gameplay and are commonly taught to novice and intermediate players. We can analyze these strategies' impact on player decision-making and move prediction by modeling them.

\textit{Controlling the Center:} Controlling the central squares (c4, c5, d4, d5, e4, e5, f4, f5) is crucial as it enhances the mobility and influence of the piece. Players can exert control directly by occupying or indirectly by targeting these squares. 

\textit{Rapid Development of Pieces:} Swift development of pieces, especially knights, bishops, and the queen, is essential in the opening phase. Early development leads to better coordination and dynamic attacking opportunities, while delayed development can result in a cramped position. 

\textit{Gaining Spatial Advantage:} Controlling more space allows greater mobility of pieces and effective execution of plans. It also prevents the opponent from ideally developing the pieces as he wants since the player controls the space. Spatial advantage often results from successful central control and piece development.

\textit{The Attack on f7:} The f7 square is a critical vulnerability in Black's position due to its proximity to the king and the lack of defense in the initial position. To exploit this weakness, White may target f7 with tactics like the Scholar's Mate or the Fried Liver Attack.

\textit{Pawn Structure and Its Strategic Implications:} Pawn structure shapes the game's strategic landscape. A solid pawn structure supports piece activity, while a weak structure can be exploited. Understanding the pawn chains and weaknesses is crucial for effective planning.

\textit{Caution with Early Queen Moves:} Premature queen moves can expose the queen to attacks and waste valuable time ("tempo" in chess terms). It also may lead to the queen being captured early in the game (which is crucial even in any level of chess). Focus on central control and piece development before bringing the queen into play.

\textit{The Role of Pawn Moves in the Opening:} Pawn moves should aim to control the center, facilitate development, or create threats. Unnecessary pawn moves can lead to strategic disadvantages.

\textit{Castling:} Castling ensures the king's safety and activates the rooks. Early castling is preferred to secure the king and bring the rooks into play. The choice between king-side and queen-side casting depends on the position and overall game plan.

\subsubsection*{Advanced Strategies Specification}
The advanced strategies we include in our model involve complex concepts, particularly material count, which is crucial for assessing the piece's value and gaining advantages.

\textit{Defending:} Prioritize piece defense to maintain or gain material advantage. Position pieces to protect one another, ensuring coordinated safety and material integrity.

\textit{Attacking and Pinning:} Aggressively target opponent's pieces to capture or pin them, especially during the opening phase. For example, if the black players positioned a knight on f6 and moved the pawn on the d column, white can position its light-square bishop on g5 and pin the black knight, exerting pressure and opening the door for further stratagems.

\textit{Trading Pieces:} Strategically exchange pieces to gain material advantages. Assess trades based on their quality to understand material imbalance and influence the game's outcome.

\subsection{Modeling Behavioral Aspects Using BP}
\label{sec:method:bp}
In the second step, we use BP to represent a chess game. Since BP is a modeling approach specifically designed to capture the behavioral aspects of complex systems, it is well-suited for representing the st- 
egic decision-making processes in chess. By decomposing the game into a set of behavioral aspects, we can model the game as a composition of strategies and heuristics that players use to make decisions. It is important to note that the BP model is not intended to replicate the full complexity of chess gameplay but rather to capture the essential behavioral patterns that drive player decisions. One of the key advantages of using BP is that it allows one to explicitly model negative aspects of the game, such as anti-scenarios, which can be used to guide the player away from undesirable moves. Here, we describe the implementation in general lines, and the reader is referred to our \href{http://...}{repository} for detailed implementation details (the link will be available upon acceptance).

We implement counters as simple scenarios that wait for any move that should be counted and then increment the counter. For example, any Pawn move increments the ``Pawn'' counter.
Strategies are more sophisticated as they are usually defined by a sequence of moves. Furthermore, the sequence may be interrupted by the enemy, and the state of the strategy may move back and forth. Thus, we implement them as a scenario that mimics this behavior and save the sequence state as an enum value.

The implementation is demonstrated in \autoref{lst:chess}. The code is written in BPjs, which is a JavaScript implementation of the BP paradigm~\cite{bar2018bpjs}. Each bthread represents a different scenario. The first bthread simulates the game, iteratively requesting that the game moves will happen in order. The second b-thread implements a simple counter for Pawn moves. The third bthread implements a basic strategy of controlling the center of the chessboard. The event set is nontrivial as it needs to take into consideration how each piece moves and whether it controls the center or not. The last bthread implements the ``Attacking and Pinning'' advanced strategy.

\begin{lstlisting}[
  style=BPjs,
  float=t,
  label={lst:chess},
  caption={A partial BP implementation.},
  numbers=none,
]
var pgn = ['e4', 'e5', 'Nf3', 'Nc6', 'Bb5', ...]
bthread('Game Simulator', function() {
  for(var m of pgn)
    request(Event('Move', m))
})

bthread('Pawn counter', function() {
  while(true) {
    e = waitFor(anyMoveES)
    if(e.piece == 'Pawn')
        request(Event('Increment', 'Pawn counter')
  }
})

// An event set that contains all moves that control the center
const ControlCenterES = ... 

bthread('Control Center', function() {
  while(true) {
    waitFor(ControlCenterES)
    request(Event('Increment', 'Control Center')
  }
}

bthread('Attacking and Pinning', function {
  var e, attack, pin
  while(true) {
    e = waitFor(anyMovesES)
    [attack, pin] = isAttackingOrPinning(e)
    if(attack)
      request(Event('Increment', 'Attacking')
    if(pin) 
      request(Event('Increment', 'Pinning')
  }
})
\end{lstlisting}

BP does not use the counters and enums. As we now elaborate, its purpose is to construct a gameplay dataset.

\subsection{Constructing the Gameplay Dataset}
The third step involves constructing a dataset that captures the state of the BP program before and after each move. Each row in this dataset represents a single move, with columns indicating the state of the game before the move, the state of the BP program after the move, and a binary label representing whether the move has been played or not. This dataset serves as the input to the machine learning model, allowing it to learn the relationship between the game state and the next move. Note that the dataset does not include the full state of the game but rather the state of the BP program, which encapsulates the essential behavioral aspects of the game. 

To create this dataset, we download games from \url{lichess.org}, filter them according to a scheme described below, and use our behavioral program to simulate each of the games. That is, move the pieces according to the recorded game steps (retrieved from the PGN) and advance the program state (i.e., the scenarios) according to the steps. At each step, we extract the program state before and after the move. We do it both for the played move and for moves that can be played but were not played. We use SMOTE to balance the unplayed and played moves.

We used the following filtering scheme for the downloaded games:

\begin{enumerate}
    \item \textit{Rapid games:} The game is played in a rapid-time format, meaning each player has 10 to 20 minutes to complete the game. This is in contrast to bullet or blitz time formats, where the moves are played more by instincts. This filter assures that the player has enough time to think about the moves and follow the playbook guidelines.
    
    \item \textit{Ranking:} In chess, players are ranked using the Elo rating system~\cite{elo}. Similar to Maia, our research focuses on intermediate players (1200--1600 Elo rating) as their gameplay tends to be more ``by the book'' and thus relatively predictable. Beginners, on the other hand, often rely more on intuition and make moves more randomly, while expert players tend to play with greater creativity and less predictable patterns. Thus we downloaded only games between two intermediate players with a rapid rating of 1200 to 1600.

    \item \textit{Complete games:} We filter out unfinished games to remove noise.
\end{enumerate}

Our work is focused on game-opening because opening strategies are easy to learn and apply and are usually used by intermediate players. Similarly, in end games, the principles just follow relatively simple routines that repeat themselves. Thus, we truncate the filtered games in the following manner:

\begin{enumerate}
    \item If both players castled, the game's opening is the series of moves until the second castle.
    \item If neither of the players has castled, or if only one of them castled in the entire game, we limit the opening part to the first 10 moves (by each player).
\end{enumerate}

\subsection{Training the Machine Learning Model}
In the final step, we train a standard machine-learning model using the constructed dataset. The model learns to predict the next move based on the current state of the game. Various machine learning algorithms can be used for this purpose, and the algorithm's choice depends on the problem's specific requirements and constraints of the problem. We compare different algorithms in \autoref{sec:eval} 

To allow the comparison with Maia, we begin with a binary classification to predict whether a human player would choose a specific move at a given level of expertise. Aside from comparing our approach to Maia, another objective is to determine how much domain-expert knowledge is required and whether only counters and basic strategies are required or also advanced strategies.

While working on binary classification, we observed that, in some cases, different players play different moves. When making decisions over the board, a human chess player does not evaluate each possible move in isolation. Instead, they assess the entire set of legal moves as a whole. From this set, they identify a subset of moves that align with their strategic goals, adhere to underlying principles, and potentially provide an advantage over the board. Among these moves, the player often deliberates, weighing the relative merits of each move based on their overarching strategy, the game's context, and the ability of a move to fulfill multiple strategic objectives simultaneously.

To better mimic this intricate decision-making process, we realized that binary classification was insufficient. Binary classification's inflexibility—treating each move as an independent decision—does not align with how human players approach the game. To address this limitation, we transitioned to a regression-based approach, which better captures a player's underlying ``debate'' between competing options. The regression framework allows us to model the relative preferences for different moves in a given position. Instead of labeling each move as a simple ``play'' or ``not play,'' regression assigns a score or percentage to each move, representing the likelihood or preference for that move to be played. This approach more closely mirrors the cognitive process of a human player, who evaluates moves not as isolated options but as part of a continuum of possibilities. Moves are evaluated collectively, considering their relationships and the strategic context of the game. This prevents moves from being treated as entirely independent entities.

Moreover, regression opens the door to uncovering deeper relationships between the strategies we rely on and the moves selected. The model can identify patterns through regression and even formulate equations linking specific strategic principles to decision-making. For example, it can reveal how heavily certain principles—such as control of the center, piece activity, or king safety—impact the preference for one move over another. This ability to derive such insights not only enhances the model's predictive accuracy but also enriches our understanding of the decision-making framework used by human players. It provides a pathway to more advanced research, as it bridges the gap between machine learning predictions and human-like strategic reasoning.

In the regression-based approach, it is necessary to convert the binary label into a numerical representation that signifies the probability of a move being played. To achieve this, we aggregated the rows in the dataset and determined the frequency of each row. Subsequently, we computed the probability of each move for every state and assigned labels to the rows based on these probabilities.

\section{Evaluation}
\label{sec:eval}

\subsection{Binary Classification}
We begin by comparing our approach with Maia in the task of binary classification. To ensure a fair comparison, we aligned the datasets and experimental setups as closely as possible. Maia's original training utilized a dataset of over 12 million games from \url{lichess.org}, providing a robust foundation for its models. For our experiments, we also accessed the same database but chose to train our models on a modest subset of only 5,000 games selected randomly. This decision was made to showcase the effectiveness of our approach. Despite using a smaller dataset and a simple machine-learning algorithm, we were able to outperform Maia. Moreover, our model is specifically tailored to track chess strategies during the opening phase. To maintain a fair comparison, we truncated the games in Maia's dataset and retrained Maia. This ensured that Maia's models were optimized for the same strategic focus as ours.

In our training process, we employed the scikit-learn library~\cite{scikit-learn} in Python to train three models: Linear Support Vector Classifier (LinearSVC), Ridge Classifier, and Logistic Regression Classifier. The default settings were used for the LinearSVC and Ridge Classifier models. For the logistic regression classifier, we conducted a comprehensive grid search algorithm with 10-fold cross-validation to optimize the hyperparameters. This process led to the selection of 4000 maximum iterations, `lbfgs' solver, L2 penalty, and $C=0.1$ as the inverse of the regularization strength.

\begin{table*}[t]
\centering
\captionsetup{justification=centerlast}
\caption{Model performance on binary classification across different Elo ratings.\\
Best results are boldfaced. The standard deviation in all cases is below 0.1\%.}
\label{tab:binary}
\renewcommand{\arraystretch}{1.5} 
\begin{adjustbox}{max width=\textwidth}
\begin{tabular}{c|c|c|c|c|c|c}
 \toprule
\multirow{2}{*}{\textbf{Strategy}} & \multirow{2}{*}{\textbf{Model}} & \multicolumn{5}{c}{ \textbf{ELO}} \\ 
 & & \textbf{1200} & \textbf{1300} & \textbf{1400} & \textbf{1500} & \textbf{1600} \\ 
\midrule
\multicolumn{2}{c|}{MAIA} & 47.3\% & 49.7\% & 51.4\% & 51.8\% & 53.1\% \\ 
\midrule
\multirow{3}{*}{\parbox{3cm}{\centering BP-Chess without\\ advanced strategies}} 
 & Linear SVC          
 & \textbf{81.80\%} 
 & 81.78\% 
 & 80.71\% 
 & 80.81\% 
 & 81.40\% 
 \\ 
 & Logistic Regression 
 & 80.95\% 
 & 81.30\% 
 & 80.21\% 
 & 80.39\% 
 & 81.01\% 
 \\ 
 & Ridge Classifier    
 & 81.27\% 
 & \textbf{81.94\%} 
 & \textbf{81.35\%} 
 & \textbf{81.50\%} 
 & \textbf{82.02\%} 
 \\
 \midrule
\multirow{3}{*}{\parbox{3cm}{\centering BP-Chess with\\ advanced strategies}} 
 & Linear SVC          
 & 82.57\% 
 & 82.29\% 
 & 80.41\% 
 & 80.10\% 
 & 80.77\% 
 \\
 & Logistic Regression 
 & 81.89\% 
 & 81.50\% 
 & 79.71\% 
 & 79.80\% 
 & 80.14\% 
 \\ 
 & Ridge Classifier    
 & \textbf{82.93\%} 
 & \textbf{83.31\%} 
 & \textbf{83.53\%} 
 & \textbf{82.53\%} 
 & \textbf{82.96\%} 
 \\
 \bottomrule
\end{tabular}
\end{adjustbox}
\end{table*}

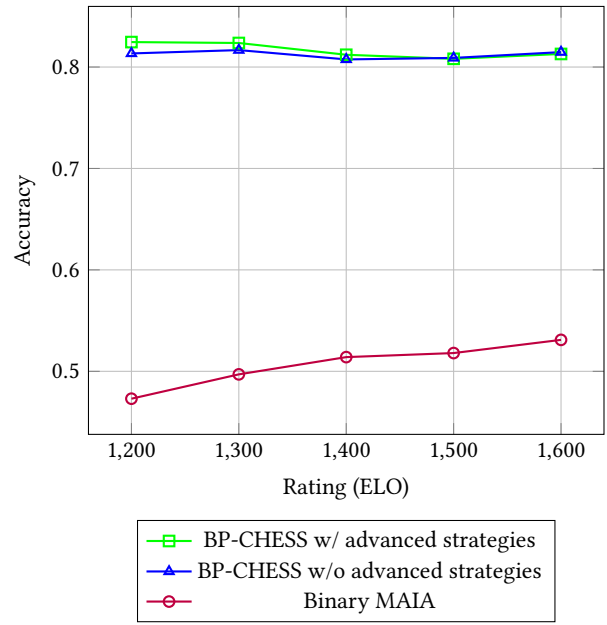
\begin{figure}
    \centering
    \begin{tikzpicture}
        \begin{axis}[
            xlabel={Rating (ELO)},
            ylabel={Accuracy},
            xtick={1200,1300,1400,1500,1600},
            ytick={0.4,0.5,0.6,0.7,0.8},
            legend style={at={(0.5,-0.2)}, anchor=north, legend columns=1},
            grid=both
            ]                
            \addplot[color=green, mark=square, thick] coordinates {
                (1200,0.8246) (1300,0.8237) (1400,0.8121) (1500,0.8081) (1600,0.8129)
            };
            \addlegendentry{BP-CHESS w/ advanced strategies}
            
            \addplot[color=blue, mark=triangle, thick] coordinates {
                (1200,0.8134) (1300,0.8167) (1400,0.8075) (1500,0.8090) (1600,0.8147)
            };
            \addlegendentry{BP-CHESS w/o advanced strategies}
            
            \addplot[color=purple, mark=o, thick] coordinates {
                (1200,0.473) (1300,0.497) (1400,0.514) (1500,0.518) (1600,0.531)
            };
            \addlegendentry{Binary MAIA}
        \end{axis}
    \end{tikzpicture}
    \caption{Binary Classification Performance}
    \label{fig:binary}
\end{figure}

\autoref{tab:binary} shows the comparison of our models with Maia for the binary classification task. Each experiment was run 10 times, and the average results were reported. The standard deviation in all cases was below 0.1\%. The results indicate that our models outperformed Maia by 27\% to 35.6\%, representing an improvement of 50\% to 75\% compared to Maia. In most cases, the Ridge classifier outperformed LinearSVC classification and logistic regression classification. Furthermore, the use of advanced strategies only marginally enhanced the predictions. We will delve into the potential reasons for this in \autoref{sec:discussion}. The results for the binary classification task are also illustrated in \autoref{fig:binary}.

\subsection{Regression}
In order to more accurately replicate the complex human decision-making process, which considers the complete set of legal moves as a cohesive unit, we shifted from binary classification to a regression-based approach. However, we were unable to match Maia's performance in the regression task.

For this task, we compared linear regression with a simple multilayer perceptron (MLP) with three fully connected layers. We also tried other ML algorithms, but linear regression outperformed all of them, so we did not include them. The MLP network had two hidden layers with 32 and 16 neurons, respectively, both utilizing ReLU activation to introduce non-linear transformations. The output layer had a single neuron and sigmoid activation. This architecture balances computational efficiency with the capacity to capture non-linear relationships in the data.

\begin{figure}
    \centering
    \begin{tikzpicture}
        \begin{axis}[
            xlabel={Rating (ELO)},
            ylabel={Mean Error},
            xtick=data,
            ymin=12,
            ymax=26,
            grid=major,
            legend style={at={(0.5,-0.2)}, anchor=north, legend columns=1},
            mark options={solid},
        ]
            
            \addplot[blue, mark=o, thick] coordinates {
                (1200,24.554) (1300,24.0848) (1400,23.9236) (1500,23.5167) (1600,23.2243)
            };
            \addlegendentry{BP-Chess w/o advanced strategies: Linear Regression}
            
            \addplot[green, mark=square*, thick] coordinates {
                (1200,24.4098) (1300,24.0561) (1400,23.7868) (1500,23.5472) (1600,23.2291)
            };
            \addlegendentry{BP-Chess with advanced strategies: Linear Regression}

            \addplot[red, mark=triangle*, thick] coordinates {
                (1200,15.7849) (1300,15.2733) (1400,15.7124) (1500,14.9050) (1600,12.5818)
            };
            \addlegendentry{BP-Chess w/o advanced strategies: MLP}         
            
            \addplot[orange, mark=diamond*, thick] coordinates {
                (1200,13.7888) (1300,13.8492) (1400,14.1364) (1500,13.2488) (1600,14.9215)
            };
            \addlegendentry{BP-Chess with advanced strategies: MLP}
            
        \end{axis}
    \end{tikzpicture}
    \caption{Regression Mean Error Performance}
    \label{fig:regression}
\end{figure}
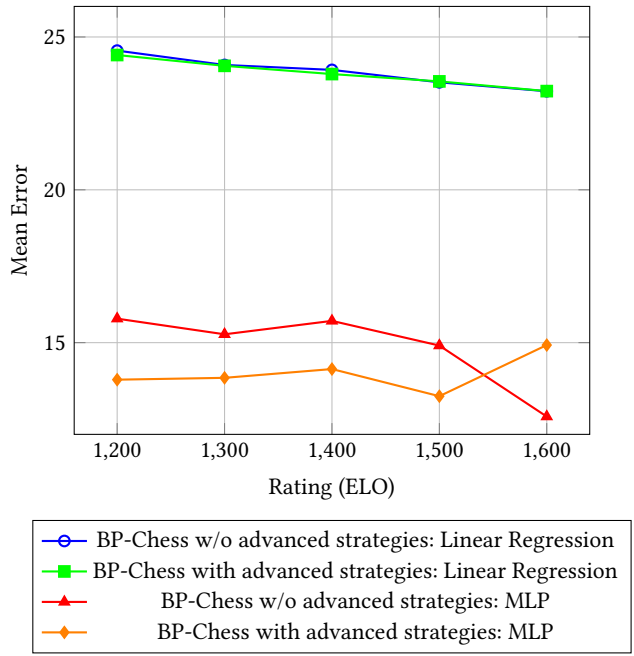

\autoref{tab:regression} shows the results for this task. Each experiment was run 10 times, and the average results were reported. The standard deviation in all cases was below 0.1\%. The results indicate that MLP is better for this task, suggesting that the human players take into consideration the complex relations between the different strategies and principles. The results are also depicted in \autoref{fig:regression}.

\begin{table*}[t]
\centering
\captionsetup{justification=centerlast }
\caption{Model performance on regression across different Elo ratings.\\
Best results are boldfaced. The standard deviation in all cases is below 0.1\%.}
\label{tab:regression}
\renewcommand{\arraystretch}{1.5} 
\begin{adjustbox}{max width=\textwidth}
\begin{tabular}{c|c|c|c|c|c|c}
\toprule
\multirow{2}{*}{\textbf{Strategy}} & \multirow{2}{*}{\textbf{Model}} & \multicolumn{5}{c}{\textbf{ELO}} \\ 
 & & \textbf{1200} & \textbf{1300} & \textbf{1400} & \textbf{1500} & \textbf{1600} \\ 
\midrule
\multirow{2}{*}{\parbox{3cm}{\centering BP-Chess without\\ advanced strategies}} 
 & Linear Regression       
 &   24.55\% 
 & 24.08\% 
 &   23.92\% 
 &  23.52\% 
 & 23.22\% 
 \\  
 & MLP  
 & \textbf{15.78\%} 
 &  \textbf{15.27\%} 
 &   \textbf{15.71\%} 
 & \textbf{14.91\%} 
 &    \textbf{12.58\%} 
 \\ 

 \midrule
 
 \multirow{2}{*}{\parbox{3cm}{\centering BP-Chess with\\ advanced strategies}} 
 & Linear Regression         
 &  24.40\% 
 &  24.01\% 
 & 23.79\% 
 &  23.55\% 
 &  23.23\% 
 \\  
 & MLP  
 &  \textbf{13.79\%} 
 &   \textbf{13.85\%} 
 &   \textbf{14.14\%} 
 & \textbf{13.25\%} 
 &  \textbf{14.92\%} 
 \\ 
 \bottomrule
\end{tabular}
\end{adjustbox}
\end{table*}

\section{Discussion}
\label{sec:discussion}
A key objective of this study was to investigate the impact of integrating complex chess strategies into machine learning models on their accuracy in predicting human moves. Our findings reveal that while augmenting advanced strategies to the fundamental behavioral model resulted in a slight enhancement in prediction accuracy, the improvement was not as significant as originally expected. This discovery prompted a deeper exploration of the correlation between strategy intricacy and move prediction accuracy.

One possible explanation for this result is the exclusive focus on the opening phase of the game. During the initial stages, chess principles like piece development, central control, and castling predominantly influence decision-making. On the other hand, advanced strategies such as piece sacrifices, positional maneuvers, and long-term tactical considerations become more prominent in the middle and endgame phases. Since our dataset was limited to the opening phase, complex strategies might not have had adequate time to materialize and substantially affect move selection. Nonetheless, their incorporation still led to a marginal yet discernible enhancement in prediction accuracy, indicating that even in the early game, human players contemplated long-term strategies, albeit less frequently.

Our study highlights the importance of knowledge representation in AI-powered chess prediction. In comparison to Maia, which relies on deep learning models trained on human games, our BP-Chess model integrates human-like strategic principles by explicitly encoding scenarios. This approach allows for the application of machine learning techniques, providing a degree of interpretability lacking in purely deep learning-based methods. While Maia is proficient in predicting moves from raw chess states, BP-Chess excels in both move prediction and in offering explanations for the reasoning behind particular moves. This capability bridges statistical predictions with human cognitive processes.


These findings highlight both the strengths and limitations of current AI approaches to modeling human decision-making in chess. Future research could explore how complex strategies influence prediction accuracy in the middle game phase, where their role is more pronounced. Additionally, refining the balance between explicit knowledge representation and deep learning techniques may lead to hybrid models that capture both human cognitive patterns and statistical learning in a more holistic manner.

\bibliographystyle{ACM-Reference-Format}
\bibliography{bibliography}
\end{document}